\documentclass[letterpaper]{article} 
\usepackage[draft]{aaai2027}
\usepackage[hyphens]{url}  
\usepackage{graphicx} 
\usepackage{natbib}  
\usepackage{caption} 
\usepackage{amsmath}
\usepackage{placeins}
\usepackage{algorithm}
\usepackage{algorithmic}

\usepackage{newfloat}
\usepackage{listings}
\DeclareCaptionStyle{ruled}{labelfont=normalfont,labelsep=colon,strut=off} 
\floatstyle{ruled}
\newfloat{listing}{tb}{lst}{}
\floatname{listing}{Listing}

\usepackage{booktabs}
\usepackage{colortbl}
\usepackage{tabularx}
\newcolumntype{Y}{>{\centering\arraybackslash}X}
\definecolor{groupgray}{gray}{0.92}
\definecolor{oursblue}{RGB}{214,239,248}
\definecolor{compgreen}{RGB}{220,248,220}

\title{LWDrive: Layer-Wise World-Model-Guided Vision-Language Model\\
Planning for Autonomous Driving}
\author{
Chen Yang\textsuperscript{\rm 1},
Yuhao Wei\textsuperscript{\rm 1},
Ze Xu\textsuperscript{\rm 1},
Ziheng Zou\textsuperscript{\rm 1},
Shuang Liang\textsuperscript{\rm 1},\\
Delin Ouyang\textsuperscript{\rm 1},
Lingfeng Qi\textsuperscript{\rm 1},
Jie Li\textsuperscript{\rm 1}\corresponding,
Guofa Li\textsuperscript{\rm 1}\corresponding
}\affiliations{ \textsuperscript{\rm 1}Chongqing University}

\begin{document}

\maketitle

\begin{abstract}
Vision-Language Models (VLMs) provide powerful semantic understanding and commonsense reasoning for End-to-End Autonomous Driving (E2E-AD) planning. However, trajectories directly generated by VLMs often encode only coarse driving intentions and remain insufficient for geometrically accurate, future-aware, and multi-view-grounded planning. To address these limitations, we develop the Layer-Wise World-Model-Guided Driving framework (\textbf{LWDrive}). LWDrive is a VLM planning framework that refines coarse trajectories through layer-wise world-model guidance. Instead of treating the VLM output as the final trajectory, LWDrive uses it as an intent-aware coarse plan, expands a diverse candidate space around it, and progressively refines the candidates through a Foresight Cascade Planner (FCP). Specifically, we introduce future-frame generation supervision to encourage the VLM to learn forward-looking scene representations, thereby injecting planning-relevant predictive dynamics into its internal hidden states. Built upon these world-model-supervised representations, FCP exploits VLM features across multiple layers and integrates historical temporal states, Action-Query representations, and current-frame multi-view Bird's-Eye-View (BEV) features to refine candidate trajectories in a coarse-to-fine manner. This design enables progressive correction of spatial positions and motion trends while grounding trajectory refinement with multi-view scene cues and preserving the high-level driving intention produced by the large model. Finally, a score head evaluates the refined candidates and selects the best trajectory as the final planning output. Experiments show that LWDrive achieves a score of 92.0 on the NAVSIM benchmark and 89.6 on NAVSIM-v2. Code and models will be made publicly available.
\end{abstract}


\section{Introduction}

End-to-End Autonomous Driving (E2E-AD) aims to directly transform sensory observations into future ego trajectories, reducing the error accumulation caused by hand-crafted modular pipelines. Recent Vision-Language Models (VLMs) and Vision-Language-Action (VLA) models further introduce semantic understanding, commonsense reasoning, and world knowledge into autonomous driving systems, offering new opportunities for handling long-tail, interaction-intensive, and rule-constrained scenarios. However, trajectory planning is not only a semantic reasoning problem. It also requires fine-grained spatial accuracy, temporal consistency, and physical feasibility under multi-view scene constraints, making direct VLM-to-trajectory decoding insufficient for reliable closed-loop planning.

\begin{figure}[t]
    \centering
    \includegraphics[width=\columnwidth]{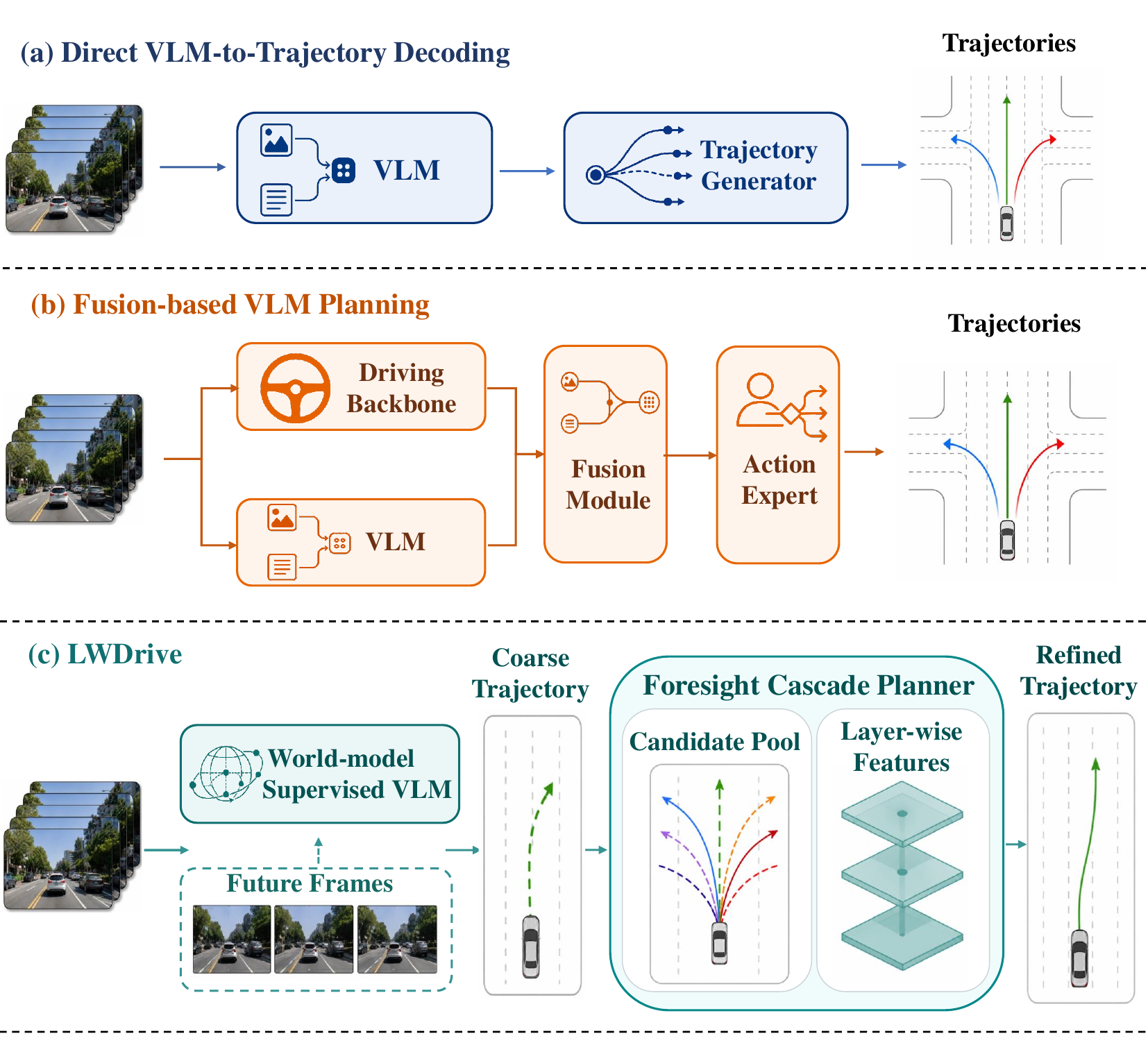}
    \caption{\textbf{Comparison of VLM planning paradigms for autonomous driving.}
    (a) Direct VLM-to-trajectory decoding lacks fine-grained correction.
    (b) Single-stage VLM/backbone fusion injects VLM semantics only once.
    (c) LWDrive performs world-model-guided coarse-to-fine refinement.}
    \label{fig:vlm_planning_paradigms}
\end{figure}

As illustrated in Figure~\ref{fig:vlm_planning_paradigms}, existing VLM-based driving planners roughly fall into two paradigms. The first directly decodes trajectories or actions from VLM representations, capturing high-level driving intentions but often lacking precise geometric and temporal constraints. The second introduces an additional driving backbone and fuses its features with VLM representations before an action expert; however, VLM semantics are usually injected only once and remain weakly coupled with subsequent trajectory correction. As a result, the rich hierarchical representations of VLMs are not fully exploited for fine-grained planning.

We argue that the role of VLMs in autonomous driving planning should not be limited to either direct trajectory generation or single-stage feature fusion. Instead, a VLM should first provide an intent-aware coarse plan, and its predictive internal representations should be used to progressively refine candidate trajectories. To enhance these representations, we introduce future-frame generation as a world-model supervision signal during training. This encourages the VLM to encode forward-looking scene dynamics without requiring future images during inference.

In this work, we develop the Layer-Wise World-Model-Guided Driving framework (\textbf{LWDrive}). It is a VLM-based planner that refines coarse trajectories into executable trajectories through layer-wise world-model guidance. LWDrive generates an intent-aware coarse trajectory with the VLM, expands it into a candidate pool, and refines the candidates through a Foresight Cascade Planner (FCP) across multiple stages. During refinement, FCP incorporates layer-wise VLM foresight features and multi-view bird's-eye-view (BEV) constraints, enabling the planner to correct spatial positions, motion trends, and scene-conditioned feasibility while preserving the high-level driving intention. This design turns VLM-based planning from one-shot trajectory generation into a progressive refinement process guided by predictive multi-layer representations.

Experiments on NAVSIM and NAVSIM-v2 demonstrate the effectiveness of LWDrive. Our method achieves 92.0 on NAVSIM and 89.6 on NAVSIM-v2, showing that world-model-supervised layer-wise VLM features provide effective guidance for coarse-to-fine autonomous driving planning.

Our contributions are summarized as follows:

\begin{itemize}
\item We introduce future-frame world-model supervision to alleviate the lack of predictive dynamics in VLM hidden states, making layer-wise representations foresight-aware and enabling effective information transfer from visual prediction to planning.
\item To address the geometric inaccuracy and weak multi-view grounding of directly decoded VLM trajectories, we develop FCP to initialize intent-aware candidates, inject selected-layer foresight through Bridge Attention, and apply BEV-grounded residual updates to progressively correct coarse trajectories.
\item We propose LWDrive, a VLM planning framework that integrates future-aware representation learning, FCP-based trajectory refinement, and score-based selection in a coarse-to-fine pipeline guided by layer-wise world-model features. Experiments show that LWDrive achieves 92.0 on NAVSIM and 89.6 on NAVSIM-v2, validating the proposed coarse-to-fine design.
\end{itemize}

\section{Related Work}

\paragraph{End-to-end Autonomous Driving.}
End-to-end autonomous driving directly maps sensory observations to future ego trajectories or control actions, reducing error accumulation and interface mismatch in modular pipelines. Early planning-oriented frameworks, such as UniAD and VAD, unify perception, prediction, mapping, and planning around the final planning objective~\cite{hu2023planning,jiang2023vad}. Recent trajectory-centric planners emphasize candidate proposal, multi-modal generation, refinement, and score-based selection. SparseDrive explores sparse scene representations, DiffusionDrive models multi-modal driving actions with truncated diffusion, while DriveSuprim, DiffRefiner, and iPad improve trajectory selection or coarse-to-fine refinement~\cite{sun2025sparsedrive,liao2025diffusiondrive,yao2026drivesuprim,yin2026diffrefiner,guo2025ipad}. Despite strong geometric and motion-level planning performance, these methods mainly rely on task-specific driving supervision, leaving high-level semantic reasoning and future-aware scene dynamics insufficiently exploited.

\paragraph{Vision-Language-Action Models for Driving.}
Vision-language and vision-language-action models introduce semantic understanding, commonsense reasoning, and instruction-following ability into autonomous driving. Early works study scene understanding, visual question answering, language-guided planning, and interpretable reasoning~\cite{sima2024drivelm,qian2024nuscenesqa,nie2024reason2drive,ma2024dolphins}. Recent planners either decode actions or trajectories from language-aligned representations~\cite{shao2024lmdrive,tian2025drivevlm,zhou2026opendrivevla,wang2025univla,zhou2025autovla,fu2025orion}, or use VLM/VLA features as semantic priors for generative planning~\cite{jiang2025diffvla,li2025recogdrive,zeng2025fsdrive,li2026sgdrive}. These methods benefit intention-level reasoning and long-tail understanding, but direct decoding can be coarse for metric planning~\cite{shao2024lmdrive,tian2025drivevlm,zhou2025autovla}, while one-shot semantic injection remains weakly coupled with later refinement~\cite{fu2025orion,jiang2025diffvla,li2025recogdrive}. LWDrive instead treats the VLM trajectory as an intent-aware coarse plan and refines candidates with layer-wise foresight features.

\paragraph{World Models for Autonomous Driving.}
World models learn predictive representations of future scene evolution and provide dense supervision for planning-oriented representation learning. Existing methods model future states through video generation~\cite{zhang2025epona,zhou2025hermes}, occupancy forecasting~\cite{zheng2024occworld}, BEV prediction and trajectory evaluation~\cite{li2025wote}, or latent dynamics for planning~\cite{li2024law,zheng2025world4drive,jia2026driveworldvla}. Prior studies show that future prediction benefits driving tasks through 4D pre-training~\cite{min2024driveworld}, latent prediction~\cite{li2024law}, BEV-based evaluation~\cite{li2025wote}, and VLA world modeling~\cite{li2025drivevlaw0,jia2026driveworldvla}. However, they often use world models as trajectory evaluators~\cite{li2025wote,zheng2025world4drive}, auxiliary objectives~\cite{min2024driveworld,li2024law,li2025drivevlaw0}, or final shared representations~\cite{zhou2025hermes,jia2026driveworldvla}. LWDrive instead makes VLM hidden states explicitly foresight-aware and exploits layer-wise predictive features for coarse-to-fine trajectory refinement.
\begin{figure*}[t]
    \centering
    \includegraphics[width=0.98\textwidth]{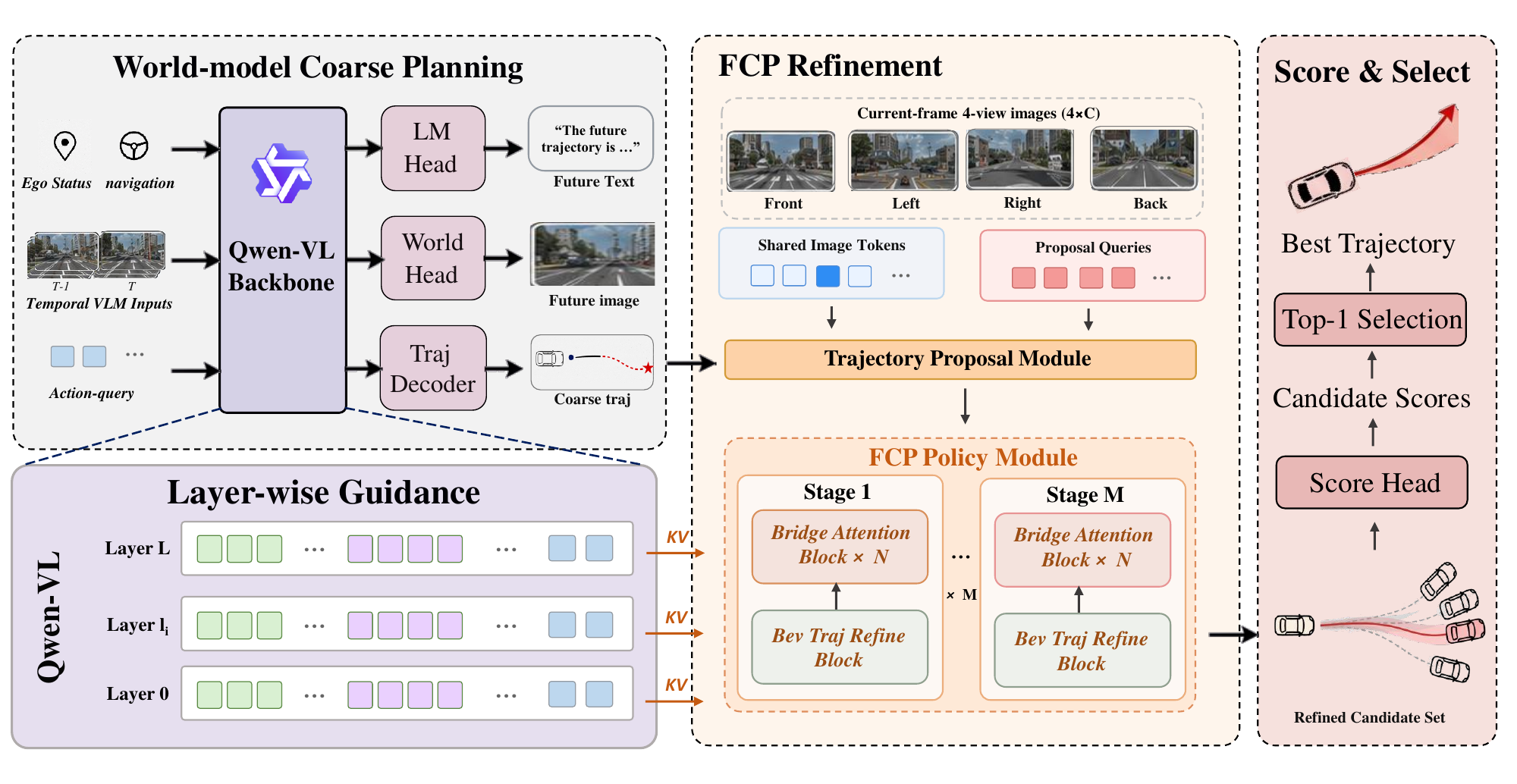}
    \caption{\textbf{Overall architecture of LWDrive.}
    LWDrive first leverages future-frame world-model supervision to guide the VLM toward predictive scene representations and an intent-aware coarse trajectory. Based on this coarse plan, it constructs a candidate trajectory pool and progressively refines it with the Foresight Cascade Planner by integrating layer-wise foresight features, temporal states, action-query memories, and multi-view BEV representations. Finally, a score head ranks the refined candidates and selects the best trajectory as the planning output.}
    \label{fig:LWDrive_overview}
\end{figure*}

\section{Methods}

\subsection{Overview}
LWDrive reformulates VLM-based autonomous driving planning from one-shot trajectory decoding into world-model-guided coarse-to-fine refinement. As shown in Figure~\ref{fig:LWDrive_overview}, the Qwen2.5-VL branch receives visual observations, ego status, navigation information, and learnable action queries. It predicts an intent-aware coarse trajectory and produces layer-wise hidden states supervised by future-frame world modeling. Future-frame supervision is applied only during training, allowing the VLM to learn predictive scene-dynamic representations while relying solely on current visual observations during inference.

Given the coarse trajectory, LWDrive constructs an expandable candidate trajectory pool and refines it through the Foresight Cascade Planner (FCP). FCP progressively injects layer-wise VLM foresight features through bridge attention and grounds the candidates with current-frame multi-view BEV features. After several refinement stages, a score head ranks all candidates and selects the final trajectory. This design allows the VLM to focus on intention-level reasoning, while FCP performs geometry-aware trajectory optimization.

\subsection{World-Model-Supervised Coarse Planning}

Before planning-oriented training, we adapt Qwen2.5-VL-3B to ego-centric driving scenes using the Impromptu dataset. This adaptation provides priors for traffic-scene understanding, navigation following, and trajectory-style response generation. We train the coarse planning branch with two objectives: future-frame world-model supervision and coarse trajectory supervision.

Given the current front-view image, ego status, navigation instruction, and learnable action-query embeddings, Qwen2.5-VL produces layer-wise hidden representations. The trajectory decoder predicts an intent-aware coarse trajectory, while the world head imposes future-frame supervision on the hidden states. Specifically, the future image $I_{t+\Delta}$ is encoded by a frozen variational autoencoder (VAE) into a clean latent target $z_{t+\Delta}$. We construct the denoising condition $c_t=[F_t^{\mathrm{V}};h_t^{\mathrm{A}}]$ by concatenating the final-layer vision hidden state and action-query hidden state. The world head predicts the clean future latent from the noisy latent $z^\tau_{t+\Delta}$:
\begin{equation}
\hat{z}_{t+\Delta}=D_{\theta}(z^\tau_{t+\Delta},c_t,\tau).
\end{equation}
The world-model loss is defined as:
\begin{equation}
\mathcal{L}_{\mathrm{wm}}=d(\hat{z}_{t+\Delta},z_{t+\Delta}).
\end{equation}

This supervision does not require future images during inference. Instead, it encourages the hidden states of Qwen2.5-VL to encode future-aware scene dynamics, making them suitable as foresight features for later candidate refinement.

In parallel, the trajectory decoder is trained to generate an intent-aware coarse trajectory that captures the high-level driving intention before candidate refinement. We denote the predicted coarse trajectory as $\hat{Y}^{0}$ and the expert trajectory as $Y^\ast$. At each future step $t$, their corresponding waypoints are denoted by $\hat{y}^{0}_{t}$ and $y^\ast_t$, respectively. The coarse trajectory loss is defined as:

\begin{equation}
\mathcal{L}_{\mathrm{traj}}
=
\frac{1}{T}
\sum_{t=1}^{T}
\mathrm{SmoothL1}(\hat{y}^{0}_{t}-y^\ast_t).
\end{equation}

The Stage-1 objective combines future-aware representation learning and coarse trajectory supervision:
\begin{equation}
\mathcal{L}_{\mathrm{stage1}}
=
\lambda_{\mathrm{wm}}\mathcal{L}_{\mathrm{wm}}
+
\lambda_{\mathrm{traj}}\mathcal{L}_{\mathrm{traj}}.
\end{equation}

During this stage, only the Qwen2.5-VL-based coarse planning branch, the world head, and the trajectory decoder are optimized. The downstream FCP and score head are not involved, which prevents the refinement modules from disturbing the world-model-supervised VLM representation.
\subsection{Foresight Cascade Planner}

After obtaining the intent-aware representation from the Qwen2.5-VL branch, FCP transforms the high-level VLM intention into executable trajectory candidates. Although the coarse trajectory captures the desired driving direction and maneuver, it may still lack fine-grained geometric accuracy, proposal-level interaction, and multi-view scene grounding. FCP therefore refines the candidate pool by progressively injecting layer-wise VLM foresight features and grounding the candidates with current-frame BEV features.

FCP first initializes an expandable candidate trajectory pool from the action-query latent representation. Instead of using the decoded coarse trajectory as the final output, we use the pooled action-query hidden state as a global intent anchor and combine it with ego-state encoding and learnable proposal embeddings for initialization:

\begin{equation}
Q^0, P^0 = \mathrm{TPM}(z_{\mathrm{AQ}}, e_t, E_{\mathrm{init}}).
\end{equation}
Here, $z_{\mathrm{AQ}}$, $e_t$, and $E_{\mathrm{init}}$ denote the pooled action-query latent, ego-state encoding, and learnable proposal embeddings, respectively. $Q^0$ denotes initialized proposal queries, and $P^0$ denotes the initial candidate pool with $N_{\mathrm{p}}$ trajectories.

To exploit hierarchical foresight information across Qwen action layers, FCP refines the proposal pool in a sparse layer-wise cascade. Let $L$ denote the total number of Qwen action layers and $M$ denote the refinement interval. FCP performs one refinement after every $M$ action layers, yielding $R$ refinement stages with the selected layer:
\begin{equation}
l_r=rM,\quad r=1,\ldots,R.
\end{equation}
This interval-based design avoids redundant updates after every layer while still exposing the planner to progressively evolved VLM representations.

\begin{figure}[!t]
\centering
\includegraphics[width=\columnwidth]{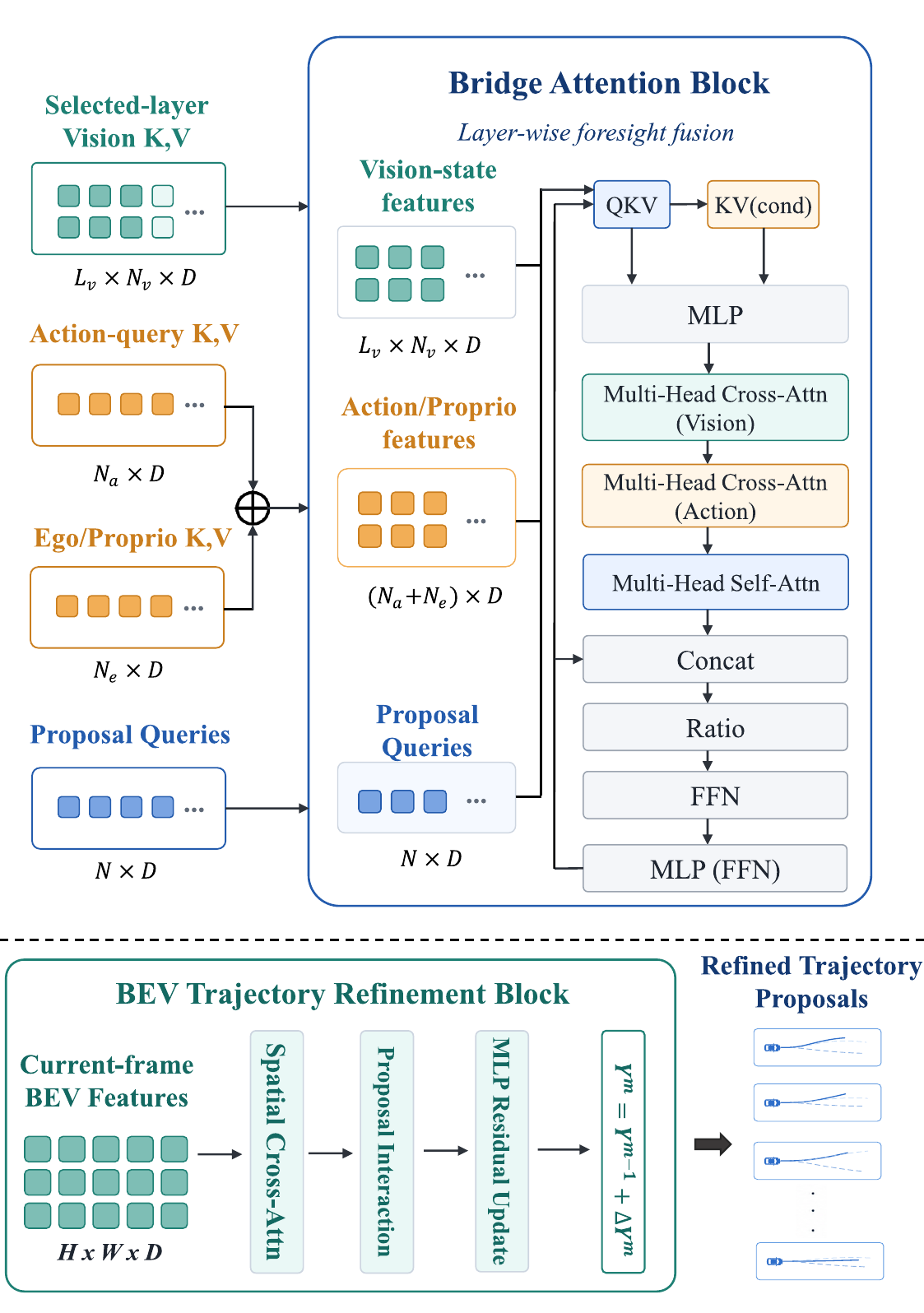}
\caption{\textbf{Foresight Cascade Planner.} Bridge Attention injects proposal interaction, action-query memory, ego-state context, and layer-wise VLM foresight features, while BEV refinement grounds the proposals with multi-view geometric cues and predicts residual updates.}
\label{fig}
\end{figure}

At the $r$-th refinement stage, we collect the selected-layer VLM features, current-frame BEV feature, and ego-state context for candidate updating as:
\begin{equation}
C_r=(H_{\mathrm{V}}^{l_r},H_{\mathrm{A}}^{l_r},B_t,s_t).
\end{equation}
The proposal queries and candidate pool are updated by:
\begin{equation}
Q^r, P^r =
\mathcal{F}r(Q^{r-1}, P^{r-1}, C_r),
\end{equation}
where $H_{\mathrm{V}}^{l_r}$ and $H_{\mathrm{A}}^{l_r}$ denote the vision/task hidden states and action-query hidden states from the selected Qwen action layer, $B_t$ denotes current-frame multi-view BEV features, and $s_t$ denotes ego-state information. Different from using only the final-layer VLM feature, this cascade schedule exposes the planner to representations with different semantic depths across refinement stages. Earlier action layers retain local visual and motion-related cues, while later layers encode higher-level intention and scene-level reasoning, allowing sparse layer-wise refinement to absorb complementary foresight information without excessive computation.

The Bridge Attention Block injects layer-wise VLM foresight into the proposal queries. It jointly attends to proposal-pool self memory, action-query and ego-state memory, and VLM foresight memory. Let $K_{\mathrm{mem}}$ and $V_{\mathrm{mem}}$ denote the concatenated multi-source keys and values from these memories. Bridge attention is computed as:
\begin{equation}
\mathrm{BA}(Q)=\mathrm{Attn}(Q,K_{\mathrm{mem}},V_{\mathrm{mem}}).
\end{equation}
This design enables each proposal to exchange information with other candidates while absorbing semantic and predictive cues from the VLM.

After bridge attention, the BEV trajectory refinement block further grounds the foresight-enhanced proposal queries with explicit multi-view geometric constraints. To achieve this, the proposal queries attend to current-frame BEV features to capture lane layout, drivable regions, road boundaries, and nearby obstacles. Each refinement stage then predicts residual trajectory updates:
\begin{equation}
\Delta Y_i^r=\mathrm{MLP}{\mathrm{ref}}(Q_i^r),
\quad
Y_i^r=Y_i^{r-1}+\Delta Y_i^r .
\end{equation}
Residual refinement stabilizes optimization and preserves the high-level driving intention inherited from the VLM branch. After $R$ stages, FCP outputs the refined candidate pool $P^R$ with $N{\mathrm{p}}$ trajectories for score-based selection.

This residual formulation prevents the refinement module from overwriting the coarse VLM intention in a single step. Instead, each stage performs a small geometry-aware correction based on proposal interaction, VLM foresight, and BEV constraints. The resulting candidate pool gradually evolves from intent-aware but coarse trajectories to executable trajectories better aligned with local scene geometry.
\subsection{Scoring and Training}

After FCP outputs the refined candidate set $P^R=\{Y_i^R\}_{i=1}^{N_{\mathrm{p}}}$, LWDrive uses a score head to evaluate all candidates and select the final planning trajectory. For each candidate, we aggregate its final proposal feature $Q_i^R$ along the temporal dimension and predict a scalar score:
\begin{equation}
S_i=\mathrm{MLP}_{\mathrm{score}}(\mathrm{Pool}(Q_i^R)).
\end{equation}
During inference, we select the highest-scoring candidate as $i^\star=\operatorname*{arg\,max}_{i} S_i$ and output $\hat{Y}=Y_{i^\star}^{R}$.

To supervise candidate selection, we compute a NAVSIM-style proposal score for each refined trajectory through non-reactive log simulation. Each candidate is tracked by a controller while surrounding agents follow their recorded trajectories. The proposal-level score is defined as:
\begin{equation}
\hat{S}_i
=
\mathrm{NC}_i \cdot \mathrm{DAC}_i \cdot
\frac{5\mathrm{EP}_i+5\mathrm{TTC}_i+2\mathrm{C}_i}{12},
\end{equation}
where $\mathrm{NC}$, $\mathrm{DAC}$, $\mathrm{EP}$, $\mathrm{TTC}$, and $\mathrm{C}$ denote No at-fault Collision, Drivable Area Compliance, Ego Progress, Time-to-Collision, and Comfort, respectively. The score head is optimized with a binary cross-entropy loss $\mathcal{L}_{\mathrm{score}}=\mathrm{BCE}(S,\hat{S})$.

LWDrive is optimized with a two-stage training strategy. In Stage 1, we train the Qwen2.5-VL-based coarse planning branch with world-model supervision and coarse trajectory supervision, while FCP and the score head are not optimized. In Stage 2, we freeze the Qwen2.5-VL backbone and action-query embeddings, and train the trajectory proposal module, FCP, and the score head.

For trajectory refinement, we supervise the candidate proposals with the expert trajectory using a Minimum-over-$N$ objective. For the $i$-th proposal at stage $r$, we define the regression loss as:
\begin{equation}
\ell_i^r=
\frac{1}{T}
\sum_{t=1}^{T}
\mathrm{SmoothL1}(y_{i,t}^{r}-y_t^\ast).
\end{equation}
To encourage progressive refinement, supervision is applied to each refinement stage:
\begin{equation}
\mathcal{L}_{\mathrm{ref}}
=
\sum_{r=1}^{R}
\beta^{R-r}
\min_i \ell_i^r ,
\end{equation}
where $Y^\ast={y_t^\ast}_{t=1}^{T}$ denotes the expert trajectory, and $\beta$ is a discount factor relaxing supervision on earlier stages.

The Stage-2 objective is:
\begin{equation}
\mathcal{L}_{\mathrm{stage2}}
=
\lambda_{\mathrm{ref}}\mathcal{L}_{\mathrm{ref}}
+
\lambda_{\mathrm{score}}\mathcal{L}_{\mathrm{score}}.
\end{equation}
This decoupled training strategy first teaches Qwen2.5-VL to learn predictive and intention-aware representations, and then trains FCP to refine and rank candidates based on frozen layer-wise foresight features.

\section{Experiments}

\begin{table*}[!t]
\centering
{\footnotesize
\setlength{\tabcolsep}{2.8pt}
\renewcommand{\arraystretch}{0.84}
\begin{tabularx}{\textwidth}{@{}l c|c|YYYYYY@{}}
\toprule
Methods & Venue & Sensors & NC$\uparrow$ & DAC$\uparrow$ & TTC$\uparrow$ & C$\uparrow$ & EP$\uparrow$ & PDMS$\uparrow$ \\
\midrule
Human driver & NeurIPS 2024 & - & 100.0 & 100.0 & 100.0 & 99.9 & 87.5 & 94.8 \\
PDM-Closed~\cite{dauner2023pdm} & PMLR 2023 & - & 94.6 & 99.8 & 89.9 & 86.9 & 99.9 & 89.1 \\
\midrule
\rowcolor{groupgray}\multicolumn{9}{l}{\emph{E2E-based Methods}} \\
UniAD~\cite{hu2023planning} & CVPR 2023 & 6$\times$C & 97.8 & 91.9 & 92.9 & \textbf{100.0} & 78.8 & 83.4 \\
LTF / TransFuser~\cite{chitta2023transfuser} & TPAMI 2022/2023 & 3$\times$C+L & 97.4 & 92.8 & 92.4 & \textbf{100.0} & 79.0 & 83.8 \\
DriveX-S~\cite{shi2025drivex} & ICCV 2025 & - & 97.5 & 94.0 & 93.0 & \textbf{100.0} & 79.7 & 84.5 \\
PRIX~\cite{wozniak2025prix} & arXiv 2025 & C & 98.1 & 96.3 & 94.1 & \textbf{100.0} & 82.3 & 87.8 \\
DiffusionDrive~\cite{liao2025diffusiondrive} & CVPR 2025 & 3$\times$C+L & 98.2 & 96.2 & 94.7 & \textbf{100.0} & 82.2 & 88.1 \\
Hydra-MDP++~\cite{li2025hydramdpplus} & CVPR 2025 & 3$\times$C+L & 98.6 & \underline{98.6} & 95.1 & \textbf{100.0} & 85.7 & 91.0 \\
iPad~\cite{guo2025ipad} & CVPR 2025 & - & 98.6 & 98.3 & 94.9 & \textbf{100.0} & \textbf{88.0} & \underline{91.7} \\
\midrule
\rowcolor{groupgray}\multicolumn{9}{l}{\emph{World-Model-based Methods}} \\
DrivingGPT~\cite{chen2025drivinggpt} & ICCV 2025 & 1$\times$C & 98.9 & 90.7 & 94.9 & 95.6 & 79.7 & 82.4 \\
World4Drive~\cite{zheng2025world4drive} & ICCV 2025 & - & 97.4 & 94.3 & 92.8 & \textbf{100.0} & 79.9 & 85.1 \\
Epona~\cite{zhang2025epona} & ICCV 2025 & 3$\times$C & 97.9 & 95.1 & 93.8 & \underline{99.9} & 80.4 & 86.2 \\
WoTE~\cite{li2025wote} & ICCV 2025 & 3$\times$C+L & 98.5 & 96.8 & 94.4 & \underline{99.9} & 81.9 & 88.3 \\
DriveWorld-VLA~\cite{jia2026driveworldvla} & CVPR 2026 & 3$\times$C & \textbf{99.1} & 98.2 & 96.1 & \textbf{100.0} & 85.9 & 91.3 \\
\midrule
\rowcolor{groupgray}\multicolumn{9}{l}{\emph{VLA-based Methods}} \\
FSDrive~\cite{zeng2025fsdrive} & NeurIPS 2025 & - & 98.2 & 93.8 & 93.3 & \underline{99.9} & 80.1 & 85.1 \\
AutoVLA~\cite{zhou2025autovla} & NeurIPS 2025 & 3$\times$C & 98.4 & 95.6 & \textbf{98.0} & \underline{99.9} & 81.9 & 89.1 \\
DriveVLA-W0~\cite{li2025drivevlaw0} & ICLR 2026 & 1$\times$C & 98.7 & \textbf{99.1} & 95.3 & 99.3 & 83.3 & 90.2 \\
ReCogDrive~\cite{li2025recogdrive} & ICLR 2026 & 3$\times$C & 97.9 & 97.3 & 94.9 & \textbf{100.0} & \underline{87.3} & 90.8 \\
SGDrive~\cite{li2026sgdrive} & CVPR 2026 & - & 98.6 & 97.8 & \underline{96.2} & \textbf{100.0} & 85.8 & 91.1 \\
\midrule
\rowcolor{oursblue}\textbf{LWDrive (Ours)} & - & \textbf{4$\times$C} & 98.8 & 98.4 & \underline{96.2} & 99.8 & \underline{87.3} & \textbf{92.0} \\
\bottomrule
\end{tabularx}
}
\caption{Comparison with state-of-the-art methods on NAVSIM. Abbreviations: NC (No at-fault Collision), DAC (Drivable Area Compliance), TTC (Time-to-Collision), C (Comfort), EP (Ego Progress), and PDMS (Predictive Driver Model Score).}
\label{tab:navsim}
\par\vspace{0.75em}

{\footnotesize
\setlength{\tabcolsep}{2.8pt}
\renewcommand{\arraystretch}{0.84}
\begin{tabularx}{\textwidth}{@{}l|YYYYYYYYYY@{}}
\toprule
Methods & NC$\uparrow$ & DAC$\uparrow$ & DDC$\uparrow$ & TLC$\uparrow$ & EP$\uparrow$ & TTC$\uparrow$ & LK$\uparrow$ & HC$\uparrow$ & EC$\uparrow$ & EPDMS$\uparrow$ \\
\midrule
Human Agent & 100.0 & 100.0 & 99.8 & 100.0 & 87.4 & 100.0 & 100.0 & 98.1 & 90.1 & 90.3 \\
\midrule
\rowcolor{groupgray}\multicolumn{11}{l}{\emph{E2E-based Methods}} \\
Ego Status MLP~\cite{li2024bevplanner} & 93.1 & 77.9 & 92.7 & 99.6 & 86.0 & 91.5 & 89.4 & \textbf{98.3} & 85.4 & 64.0 \\
TransFuser~\cite{chitta2023transfuser} & 96.9 & 89.9 & 97.8 & \underline{99.7} & 87.1 & 95.4 & 92.7 & \textbf{98.3} & 87.2 & 76.7 \\
Hydra-MDP++~\cite{li2025hydramdpplus} & 97.2 & 97.5 & 99.4 & 99.6 & 83.1 & 96.5 & 94.4 & \underline{98.2} & 70.9 & 81.4 \\
DriveSuprim~\cite{yao2026drivesuprim} & 97.5 & 96.5 & 99.4 & 99.6 & \underline{88.4} & 96.6 & 95.5 & \textbf{98.3} & 77.0 & 83.1 \\
ARTEMIS~\cite{feng2025artemis} & 98.3 & 95.1 & 98.6 & \textbf{99.8} & 81.5 & 97.4 & 96.5 & \textbf{98.3} & \textbf{98.3} & 83.1 \\
PRIX~\cite{wozniak2025prix} & 98.0 & 95.6 & \underline{99.5} & \textbf{99.8} & 87.4 & 97.2 & \textbf{97.1} & \textbf{98.3} & 87.6 & 84.2 \\
DiffusionDrive~\cite{liao2025diffusiondrive} & 98.2 & 95.9 & 99.4 & \textbf{99.8} & 87.5 & 97.3 & 96.8 & \textbf{98.3} & \underline{87.7} & 84.5 \\
\midrule
\rowcolor{groupgray}\multicolumn{11}{l}{\emph{VLA / World-Model-based Methods}} \\
DriveVLA-W0~\cite{li2025drivevlaw0} & 98.5 & \textbf{99.1} & 98.0 & \underline{99.7} & 86.4 & \underline{98.1} & 93.2 & 97.9 & 58.9 & 86.1 \\
DriveWorld-VLA~\cite{jia2026driveworldvla} & \underline{98.6} & \textbf{99.1} & \textbf{99.6} & \textbf{99.8} & 87.4 & 97.9 & \underline{97.0} & 97.8 & 78.6 & \underline{86.8} \\
\midrule
\rowcolor{oursblue}\textbf{LWDrive (Ours)} & \textbf{98.8} & \underline{98.4} & 99.0 & \underline{99.7} & \textbf{90.3} & \textbf{98.6} & 96.3 & 97.9 & 73.3 & \textbf{89.6} \\
\bottomrule
\end{tabularx}
}
\caption{Comparison with state-of-the-art methods on NAVSIM-v2. Abbreviations: NC (No at-fault Collision), DAC (Drivable Area Compliance), DDC (Driving Direction Compliance), TLC (Traffic Light Compliance), EP (Ego Progress), TTC (Time-to-Collision), LK (Lane Keeping), HC (Human Comfort), EC (Extended Comfort), and EPDMS (Extended Predictive Driver Model Score).}
\label{tab:navsimv2}
\end{table*}

\begin{figure*}[!t]
\centering
\includegraphics[width=0.98\textwidth]{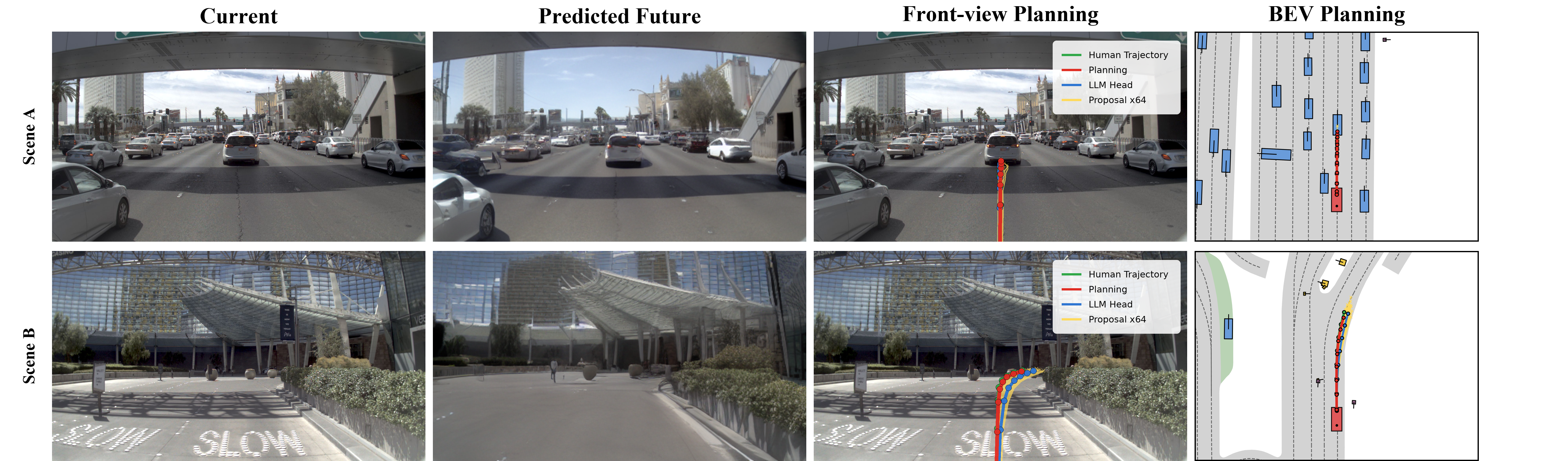}
\caption{\textbf{Qualitative visualization of LWDrive.} For each scene, we show the current front-view image, the future frame predicted by the world-model head, the trajectory planning result projected onto the front-view image, and the corresponding BEV trajectory planning result.}
\label{fig:visualization_two_scenarios}
\end{figure*}

\subsection{Experimental Setup}

\paragraph{Datasets and Metrics.}
We conduct experiments on three planning benchmarks: NAVSIM~\cite{dauner2024navsim}, NAVSIM-v2~\cite{cao2025navsimv2}, and nuScenes~\cite{caesar2020nuscenes}. NAVSIM and NAVSIM-v2 support closed-loop planning evaluation, while nuScenes is used for open-loop evaluation with ST-P3. For NAVSIM, we report the Predictive Driver Model Score (PDMS), which combines No at-fault Collision (NC), Drivable Area Compliance (DAC), Time-to-Collision (TTC), Comfort (C), and Ego Progress (EP). For NAVSIM-v2, we report the Extended PDMS (EPDMS), which further considers rule- and lane-level driving behaviors. For nuScenes, we report average $L_2$ displacement error and collision rate.
\paragraph{Implementation.}
LWDrive follows a two-branch implementation. The VLM branch is initialized from Qwen2.5-VL-3B~\cite{bai2025qwen25vl} and processes a front-view image resized to $280\times504$. For trajectory refinement, FCP adopts a ResNet-34~\cite{he2016resnet} encoder over four camera views to extract BEV features. The model contains 36 Qwen action layers and 6 trajectory refinement modules. Training uses AdamW and proceeds in two stages. In Stage 1, we train the coarse planning branch for 8,000 steps with learning rate $2\times10^{-4}$, global batch size 48, and loss weights 15.0/1.0 for world-model and trajectory supervision, requiring about 100 hours. In Stage 2, we freeze the Qwen2.5-VL backbone and action-query embeddings, and optimize FCP and the score head for 12 epochs with learning rate $3\times10^{-5}$ and global batch size 64, requiring about 80 hours.
\subsection{Main Results}

\paragraph{NAVSIMv1.}
We first evaluate LWDrive on NAVSIM, with the results reported in Table~\ref{tab:navsim}. LWDrive achieves the best PDMS of 92.0 among the compared methods, surpassing strong proposal-based and VLA-based planners such as iPad, DriveWorld-VLA, and SGDrive. It also attains competitive safety scores with 98.8 NC and 96.2 TTC, while reaching a high EP of 87.3. This indicates that the proposed world-model-guided coarse-to-fine refinement improves driving progress without sacrificing safety.

\paragraph{NAVSIMv2.}
We further evaluate LWDrive under the extended NAVSIM-v2 protocol, with results shown in Table~\ref{tab:navsimv2}. LWDrive achieves an EPDMS of 89.6, outperforming all compared methods, including DriveVLA-W0 and DriveWorld-VLA. The advantage is mainly reflected in progress and interaction safety: LWDrive obtains the best EP and TTC scores, 90.3 and 98.6, while maintaining strong rule-related metrics such as NC, DAC, DDC, TLC, and LK. These results show that layer-wise foresight features and BEV-based refinement help the planner generate efficient trajectories consistent with local driving constraints.

\paragraph{nuScenes.}
We also evaluate LWDrive on the nuScenes ST-P3 benchmark, with open-loop planning results reported in Table~\ref{tab:nuscenes}. LWDrive achieves the lowest average $L_2$ error of 0.37 m and a competitive collision rate of 0.16\%. Compared with prior autoregressive, non-autoregressive, and VLA-based planners, its strong trajectory accuracy shows that LWDrive can transfer beyond NAVSIM-style scoring while maintaining reliable motion prediction. This suggests that world-model-supervised foresight representations provide useful motion cues for both closed-loop-style evaluation and open-loop trajectory prediction, especially in complex, motion-intensive, and safety-critical driving scenarios.

\paragraph{Overall.}
LWDrive improves closed-loop-style and open-loop planning by refining VLM priors into BEV-grounded trajectories. NAVSIM/NAVSIM-v2 gains reflect progress, safety, and rule compliance, while nuScenes shows transfer. PDMS/EPDMS follow official evaluator: scores are scenario-averaged, while sub-metrics are independently averaged and cannot reconstruct aggregate scores.
\par\noindent
\begin{minipage}{\columnwidth}
\centering
\setcounter{table}{2}
\footnotesize
\setlength{\tabcolsep}{3.2pt}
\renewcommand{\arraystretch}{0.88}
\begin{tabularx}{\columnwidth}{@{}Xcc@{}}
\toprule
\textbf{Method} & \textbf{$L_2$ (m)$\downarrow$} & \textbf{CR (\%)$\downarrow$} \\
\midrule
\rowcolor{groupgray}\multicolumn{3}{l}{\emph{Non-Autoregressive Methods}} \\
ST-P3~\cite{hu2022stp3} & 2.11 & 0.71 \\
VAD~\cite{jiang2023vad} & 1.25 & 1.09 \\
Ego-MLP~\cite{li2024bevplanner} & 0.78 & 0.38 \\
UniAD~\cite{hu2023planning} & 0.69 & \underline{0.12} \\
InsightDrive~\cite{song2025insightdrive} & 0.44 & 0.15 \\
BEV-Planner~\cite{li2024bevplanner} & 0.55 & 0.59 \\
\midrule
\rowcolor{groupgray}\multicolumn{3}{l}{\emph{Autoregressive Methods}} \\
DriveVLM~\cite{tian2025drivevlm} & \underline{0.40} & 0.27 \\
GPT-Driver~\cite{mao2023gptdriver} & 0.44 & 0.17 \\
OccWorld~\cite{zheng2024occworld} & 0.77 & 0.32 \\
Doe-1~\cite{zheng2024doe1} & 0.70 & 0.21 \\
RDA-Driver~\cite{huang2024rdadriver} & \underline{0.40} & \textbf{0.10} \\
OpenEMMA~\cite{xing2025openemma} & 2.81 & - \\
DME-Driver~\cite{han2024dmedriver} & 0.98 & 0.29 \\
OmniDrive~\cite{wang2024omnidrive} & 0.84 & 0.94 \\
AutoVLA (action only)~\cite{zhou2025autovla} & 0.43 & 0.19 \\
AutoVLA (w/ CoT)~\cite{zhou2025autovla} & 0.48 & 0.13 \\
\midrule
\rowcolor{oursblue}\textbf{LWDrive (Ours)} & \textbf{0.37} & 0.16 \\
\bottomrule
\end{tabularx}
\captionof{table}{Comparison with selected related methods on the nuScenes ST-P3 benchmark. Lower is better.}
\label{tab:nuscenes}
\end{minipage}
\par\smallskip
\subsection{Visualization}

Figure~\ref{fig:visualization_two_scenarios} shows that the predicted future frames provide plausible scene-evolution cues, while the front-view and BEV planning results remain consistent with the driving intention and local road geometry. These qualitative examples suggest that world-model-supervised foresight features and multi-view BEV constraints help refine coarse VLM plans into smooth and feasible trajectories.

\subsection{Ablation Study}

We conduct ablation studies on NAVSIM to analyze the main design choices of LWDrive under a unified evaluation setting, including world-model-supervised coarse planning, layer-wise foresight feature usage, Bridge Attention, and BEV trajectory refinement. The first two rows compare direct VLM decoding with world-model-supervised coarse planning, while the remaining variants examine how different refinement components affect planning performance.

As shown in Table~\ref{tab:ablation}, using the VLM-decoded trajectory gives a PDMS of 84.5, indicating that the VLM output provides a high-level planning prior rather than an executable trajectory. Adding world-model supervision improves PDMS to 86.3, and LWDrive raises it to 92.0, validating coarse-to-fine refinement. Removing BEV refinement drops PDMS to 89.3, while removing Bridge Attention gives 90.0, showing that geometric grounding and foresight injection are both important. The final-layer variant reaches 91.8, below the layer-wise design, suggesting that useful cues are distributed across Qwen action layers. These results verify that predictive VLM representations, layer-wise feature injection, Bridge Attention, and BEV-grounded refinement jointly contribute to planning performance.

\begin{center}
\begin{minipage}{\columnwidth}
\centering
\footnotesize
\setlength{\tabcolsep}{1.1pt}
\renewcommand{\arraystretch}{0.84}
\begin{tabular*}{\columnwidth}{@{}c@{}c@{}c@{}c|@{\extracolsep{\fill}}cccccc@{}}
\toprule
WM & LW & BA & BR & NC$\uparrow$ & DAC$\uparrow$ & EP$\uparrow$ & TTC$\uparrow$ & C$\uparrow$ & PDMS$\uparrow$ \\
\midrule
$\times$ & $\times$ & $\times$ & $\times$ & 98.2 & 92.7 & 79.1 & 93.9 & 100.0 & 84.5 \\
$\surd$ & $\times$ & $\times$ & $\times$ & 98.4 & 94.7 & 81.5 & 94.7 & 99.9 & 86.3 \\
$\surd$ & $\surd$ & $\surd$ & $\times$ & 98.9 & 97.2 & 83.9 & 95.2 & 99.8 & 89.3 \\
$\surd$ & $\surd$ & $\times$ & $\surd$ & 98.7 & 97.7 & 84.6 & 95.1 & 99.7 & 90.0 \\
$\surd$ & Final & $\surd$ & $\surd$ & 98.7 & 98.0 & 88.1 & 95.7 & 99.7 & 91.8 \\
\midrule
\cellcolor{compgreen}$\surd$ & \cellcolor{compgreen}$\surd$ & \cellcolor{compgreen}$\surd$ & \cellcolor{compgreen}$\surd$ & \textbf{98.8} & \textbf{98.4} & 87.3 & \textbf{96.2} & 99.8 & \textbf{92.0} \\
\bottomrule
\end{tabular*}
\captionof{table}{Ablation on NAVSIM. WM, LW, BA, and BR denote world-model supervision, layer-wise foresight, Bridge Attention, and BEV refinement, respectively. ``Final'' in LW denotes using only the final-layer VLM feature instead of layer-wise features.}
\label{tab:ablation}
\end{minipage}
\end{center}

\section{Conclusion}

We presented LWDrive, a layer-wise world-model-guided coarse-to-fine planning framework for VLM-based autonomous driving. LWDrive treats the VLM output as an intent-aware coarse plan, expands it into a candidate pool, and progressively refines the candidates with layer-wise foresight features and multi-view BEV cues. Experiments on NAVSIM, NAVSIM-v2, and nuScenes demonstrate that combining predictive VLM representations with explicit trajectory refinement and selection improves the accuracy and feasibility of autonomous driving planning.

\bibliography{aaai2027}

\end{document}